\documentclass{article}
\usepackage{spconf,amsmath,bm,graphicx}
\usepackage{cite,url}
\usepackage{hyperref}
\usepackage{nicefrac}
\usepackage{graphicx}
\usepackage{color}
\usepackage{booktabs}
\usepackage[inline]{enumitem}
\usepackage[group-four-digits]{siunitx}
\newcommand{\tS}[1]{\textsuperscript{#1}}

\newcommand{\y}{\bm{y}}
\newcommand{\x}{\bm{x}}
\renewcommand{\o}{\bm{o}}

\newcommand{\h}{\bm{h}}
\newcommand{\W}{\bm{W}\!}

\newcommand{\Y}{\mathcal{Y}}
\renewcommand{\L}{\mathcal{L}}

\title{A Large-Scale Study of Language Models\\for Chord Prediction}

\name{Filip Korzeniowski$^{\star}$  \qquad David R. W. Sears$^{\dagger}$ \qquad Gerhard Widmer$^{\star}$
\thanks{This work is supported by the European Research Council (ERC) under the EU's Horizon 2020 Framework Programme (ERC Grant Agreement number 670035, project ``Con Espressione'').}}
\address{$^{\star}$ Department of Computational Perception, Johannes Kepler University, Linz, Austria\\
         $^{\dagger}$ College of Visual \& Performing Arts, Texas Tech University, Lubbock, USA}

\begin{document}
\ninept{}
\maketitle
\begin{abstract}
We conduct a large-scale study of language models for chord prediction.
Specifically, we compare $N$-gram models to various flavours of recurrent
neural networks on a comprehensive dataset comprising all publicly available
datasets of annotated chords known to us. This large amount of data allows us
to systematically explore hyper-parameter settings for the recurrent neural
networks---a crucial step in achieving good results with this model class. Our
results show not only a quantitative difference between the models, but also a
qualitative one: in contrast to static $N$-gram models, certain RNN
configurations adapt to the songs at test time. This finding constitutes a further
step towards the development of chord recognition systems that are more aware
of local musical context than what was previously possible.
\end{abstract}


\section{Introduction}

Chord recognition is a long-standing topic in the music information retrieval
community. Steady advances notwithstanding, results have seemed to stagnate in
recent years. The study in~\cite{humphrey_four_2015} offers possible
explanations, including invalid harmonic assumptions, limitations of evaluation
measures, conflicting problem definitions, and the subjectivity inherent to
this task. All of these criticisms are certainly valid, and in our view, the
latter two bear the greatest potential to hamper further improvements. However,
at present there is still a large performance gap between human annotators
(measured via inter-annotator agreement) and automatic
systems~\cite{humphrey_four_2015}, although they are subjected to the same
constraints.

We can divide a chord recognition system into two parts: \emph{hearing}
(acoustic modelling, feature extraction) and \emph{understanding} (combining
observations, finding a meaningful chord sequence). Research has focused mainly
on the former by finding better features and acoustic models
\cite{humphrey_rethinking_2012,korzeniowski_feature_2016,ueda_hmmbased_2010},
while only a few works have explored improvements of the latter using
temporal models
\cite{boulanger-lewandowski_audio_2013,mauch_approximate_2010,pauwels_combining_2014}.
This trend was reinforced through the insight that existing temporal models do
little but smooth the noisy predictions of the acoustic model over time, i.e.\
they mainly model the chord's duration
\cite{chen_chord_2012,cho_relative_2014}.

Thus, we suggest another possible explanation for the performance
gap: current state-of-the-art chord recognition systems lack a meaningful
understanding of harmony and its development in music. As shown in
\cite{korzeniowski_futility_2017}, such an understanding can only derive from
sequential models that operate at higher temporal levels than those based on audio
frames. 

The present work addresses this issue by exploring the
capabilities of automatically learned chord language models to predict chord
sequences. To our knowledge, this paper constitutes the first attempt to
systematically find and evaluate such models. We have seen work on optimising
low-level and/or Markovian \emph{temporal} models in e.g.\
\cite{cho_relative_2014}, and work on low-level non-Markovian models in e.g.\
\cite{boulanger-lewandowski_audio_2013}, but as argued in
\cite{korzeniowski_futility_2017}, at the level of audio frames, there is
little to improve upon the current state-of-the-art.

Our work compares recurrent neural networks (RNNs) with higher-order $N$-gram
models. We evaluate these models independently based on how well they predict
chord sequences. The question on how to integrate them into a complete chord
recognition system is left for future work.

\section{Data}\label{sec:data}

We compiled a comprehensive set of chord annotations to perform a large-scale
evaluation of different language models for chord prediction. To our knowledge,
our compound dataset consists of all time-aligned chord annotations that are
publicly available.
Table~\ref{tab:data} provides detailed information about the data.  In total,
we have \num{1841} songs from a variety of genres and decades, with a focus on
pop/rock between 1950 and 2000. After we remove duplicate songs and merge
consecutive identical annotations, the dataset consists of \num{1766} songs
containing \num{161796} unique chord annotations.

\begin{table}[]
\centering
\begin{tabular}{@{}lcrr@{}}
\toprule
\textbf{Name}    & \textbf{Ref.}                                 & \textbf{No. Pieces} & \textbf{No. Chords} \\ \midrule
Beatles          &~\cite{harte_automatic_2010}                   & 180                 & \num{12646}               \\
Jay Chou         &~\cite{deng_automatic_2016}                    & 29                  & \num{3356}                \\
McGill Billboard &~\cite{burgoyne_expert_2011}                   & 742                 & \num{70197}              \\
Queen            &~\cite{mauch_omras2_2009}                      & 20                  & \num{2265}                 \\
Robbie Williams  &~\cite{digiorgi_automatic_2013}                & 65                  & \num{6513}                \\
Rock             &~\cite{declercq_corpus_2011}                   & 201                 & \num{18343}               \\
RWC              &~\cite{goto_rwc_2002}                          & 100                 & \num{12726}               \\
US Pop 2002      &~\cite{ellis_uspop2002_2003}                   & 195                 & \num{23309}               \\
Weimar Jazz      &~\cite{thejazzomatresearchproject_weimar_2016} & 291                 & \num{18179}               \\
Zweieck          &~\cite{mauch_omras2_2009}                      & 18                  & \num{1822}                \\ \midrule
Total            &                                               & \num{1841}          & \num{169356}              \\
\textbf{Unique}  & \textbf{}                                     & \textbf{\num{1766}}       & \textbf{\num{161796}}     \\ \bottomrule
\end{tabular}
\caption{Individual datasets used to create the compound dataset in this study. For the Rock corpus,
we used the annotations by Temperley rather than those by De Clerq.} 
\label{tab:data}
\end{table}

\newpage
The dataset is miniscule compared to those available for natural language
processing: e.g., the NYT section of the English Gigaword dataset (of which
only a subset of 6.4 million words is used in~\cite{mikolov_recurrent_2010} for
training language models) consists of 37 million words. 
Therefore,
we leverage domain knowledge to generate additional, semi-artificial chord
sequences. Assuming that chord progressions are independent of musical key (as
in Roman numeral analysis), and that musical keys are uniformly distributed (a
simplifying assumption), we transpose each chord sequence to all possible keys
during training. This step increases the amount of training sequences by a factor of
12; however, the artificially created data are highly correlated with the
existing data, and thus are not equivalent to truly having 12 times as much data
available. We will refer to this process as data augmentation in the remainder
of this paper.

We focus on the major/minor chord vocabulary, and follow the standard mapping
as described in~\cite{cho_relative_2014}: chords with a minor 3\tS{rd} in the
first interval are considered minor, the rest major. This mapping results in a
vocabulary of size 25 ($12$ root notes $\times \{\text{maj}, \text{min}\}$ 
and the ``no-chord'' class). Bearing in mind the reduced
vocabulary and the more repetitive nature of chord progressions compared to
spoken language, we feel that the size of this dataset after data augmentation
is appropriate for training and evaluating models for chord prediction.

\section{Experimental Setup}\label{sec:experimental_setup}

Our experiments evaluate how well various models predict chords. This
amounts to measuring the average probability the model assigns to the (correct)
upcoming chord symbol, given the ones it has already observed in a sequence.
More formally, given a sequence of chords $\y = (y_1,\ldots,y_K)$, the model
$M$ yields $P_M(y_k \mid y_1^{k-1})$ for each $k$. (We denote
$y_1, \ldots, y_{k-1}$ as $y_1^{k-1}$.) The probability of the
complete chord sequence can then be computed as
\begin{align}
P_M\left(\y\right) = P_M(y_1) \cdot \Pi_{k=2}^{K} P_M\left(y_k \mid y_1^{k-1}\right).
\label{eq:seq_prod}
\end{align}

The higher $P_M(\y)$, the better $M$ models $\y$. To measure how well $M$
models all chord sequences in a dataset $\Y$, we next compute the average log-probability
assigned to its sequences:
\begin{align}
  \L(M,\Y) = \frac{1}{N_\Y} \sum_{\y \in \Y} \log\left(P_M\left(\y\right)\right),
  \label{eq:logp}
\end{align}
where $N_\Y$ is the total number of chord symbols in the dataset. This equation corresponds
to the negative cross-entropy between the model's distribution and the data
distribution. 

All models are trained and tested on the compound dataset. We use 20\% of the
data for testing, and 80\% for training. 15\% of the training data is held out
for validation. All splits are stratified by dataset.

\section{$N$-Gram Language Models}\label{sec:n-gram_language_models}

$N$-gram language models are Markovian probabilistic models that assume a
fixed-length history (of length $N - 1$) in order to predict the next
symbol. Hence, they assume $P_M(y_k \mid y_1^{k-1}) = P_M(y_k \mid y_{k - N + 1}^{k - 1})$.
This fixed-length history allows the probabilities to be stored in a table, with
its entries computed using maximum-likelihood estimation (i.e., by counting
occurrences in the training set).

With larger $N$, the sparsity of the probability table increases exponentially due
to the finite number of $N$-grams in the training set. We solve
this problem using Lidstone smoothing, which adds a pseudo-count $\alpha$ to each
possible $N$-gram. We determine the value of $\alpha$ for each model using the
validation set.

\subsection{Model Selection}

We find the best $N$-gram model by selecting the one with best results on the
validation set. To this end, we evaluate models with
$N \in \{1, 2, 3, 4, 5, 6\}$, with and without data augmentation. $N=1$
corresponds to a model that predicts chords just by their frequency in the
training data, $N=2$ to a model that could be deployed in a simple first-order
Hidden Markov Model.

\begin{figure}
  \centering 
  \includegraphics{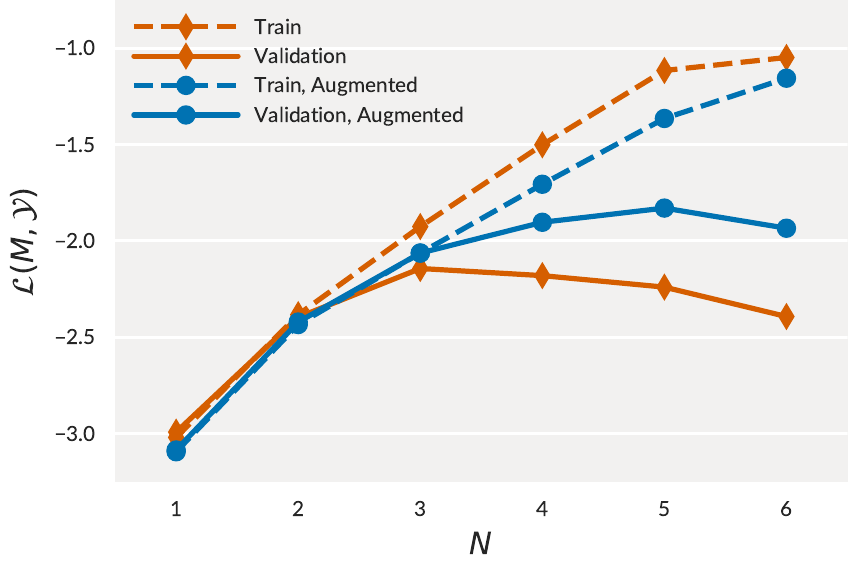}
  \caption{Average log-probability of all evaluated $N$-gram models.}\label{fig:ngram_results}
\end{figure}

Figure~\ref{fig:ngram_results} presents the results of the models on the
training and validation sets. With data augmentation, the best result is
achieved by a $5$-gram model ($\alpha=0.3$).

\section{Recurrent Neural Language Models}\label{sec:recurrent_neural_language_models}

Recurrent Neural Networks (RNNs, see~\cite{pascanu_how_2014} for an overview)
are powerful models designed for sequential modelling tasks.  In their simplest
form, RNNs transform input sequences $\x_1^K$ to an output sequence $\o_1^K$
through a non-linear projection into a hidden layer $\h_1^K$, parameterised by
weight matrices $\W_{hx}$, $\W_{hh}$ and $\W_{oh}$:
\begin{align}
  \h_k &= \sigma_h\left(\W_{hx}\x_k + \W_{hh}\h_{k-1}\right) \label{eq:hidden_layer}\\
  \o_k &= \sigma_o\left(\W_{oh}\h_k\right) \label{eq:output_layer},
\end{align}
where $\sigma_h$ and $\sigma_o$ are the activation functions for the hidden layer
(e.g.\ the sigmoid function), and the output layer (e.g.\ the softmax),
respectively. We left out bias terms for simplicity.

\begin{figure}
  \centering \includegraphics[width=.9\columnwidth]{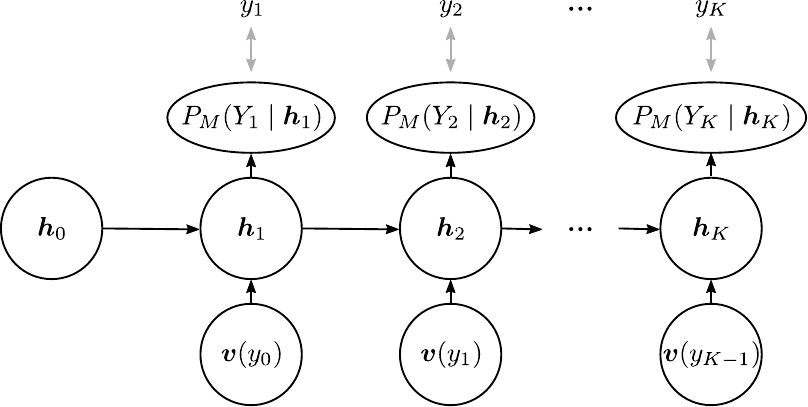}
  \caption{A simple RNN-based language model. We can easily stack more
          recurrent hidden layers or add skip-connections between the input and
          each hidden layer or the output.}\label{fig:rnn_lang}
\end{figure}

Their use as language models was proposed in~\cite{mikolov_recurrent_2010}. For
this purpose, the input at each time step $k$ is a vector representation of
the preceding symbol $y_{k-1}$. We call this the \emph{chord embedding}, and denote
it $\bm{v}(y_{k-1})$. In the simplest case, this is a one-hot vector.
The network's output $\o_k$ is then interpreted as the conditional probability
over the next chord symbol $P_M\left(Y_k \mid y_1^{k-1}\right)$. During
training, the categorical cross-entropy between the $\o_k$ and the true chord
symbol is minimised by adapting the weight matrices in
Eq.~\ref{eq:hidden_layer} and~\ref{eq:output_layer} using stochastic gradient
descent and back-propagation through time. Figure~\ref{fig:rnn_lang} provides an
overview of the model structure.

Each output $\o_k$ depends on all previous inputs $y_1^{k-1}$ through the
recurrent connection in the hidden layer. This allows the network to consider
all previous chords when computing $P_M(y_k \mid y_1^{k-1})$. In practice,
this capacity is limited because of the limited size of the hidden layer, and
the fragile learning procedure of back-propagation through time, which faces the
well-known problems of exploding and vanishing gradients.

\subsection{Chord Embeddings}\label{sec:chord_embeddings}

As mentioned earlier, we need to represent chord classes as vectors to use them
in the RNN language model framework. In this work, we explore three
possibilities: 
\begin{enumerate*}[label=\arabic*)]
\item using the one-hot encoding of the class, 
\item using a fixed-length vector that is learned jointly with the language 
      model, and 
\item learning an embedding using the word2vec skip-gram model
\cite{mikolov_efficient_2013} before training the language model itself.
In this case, the chord embeddings are optimised to predict the neighbouring
chords.
\end{enumerate*}

\subsection{Model Selection}\label{sec:model_selection}

RNNs have more hyper-parameters compared to $N$-gram models: we can set the
number of hidden layers, the size of each hidden layer, the type and
dimensionality of the input chord embedding, the activation function for each
hidden layer, whether to use skip-connections, and finally, we can decide
to use simple RNNs or more advanced hidden layer structures such as Long
Short-Term Memory (LSTM)~\cite{hochreiter_long_1997} or Gated Recurrent Units
(GRU)~\cite{cho_properties_2014}. Additionally, the training procedure
has its own hyper-parameters, such as
learning rate or mini-batch size. Table~\ref{tab:hypers}
presents the hyper-parameter space we sampled from.

\begin{table}[]
\small
\centering
\begin{tabular}{@{}ll@{}}
\toprule
\textbf{Hyper Parameter} & \textbf{Sample Space}                                                                         \\ \midrule
Embedding Size           & $\{4, 8, 16, 24\}$                                                                            \\
Embedding Type           & $\{\text{one-hot}, \text{word2vec}, \text{learned}\}$                                         \\
No. Hidden Layers        & $N_h \in \{1, 2, 3, 4, 5\}$                                                                   \\
Hidden Layer Size        & $D_h \in \{128, 256, 512, 1024\}$                                                                     \\
Skip Connections         & $\{\text{yes}, \text{no}\}$                                                                   \\
Learning Rate            & $lr = \begin{cases} \{0.001, 0.0005\} & N_h \le 3\\ \{0.0005, 0.00025\} & \text{else} \end{cases}$ \\
Learning Rate (GRU)      & $lr = \begin{cases}\{0.005, 0.001\} & N_h \le 3\\\{0.001, 0.0005\} & \text{else}\end{cases}$       \\ \bottomrule
\end{tabular}
\caption{Hyper-parameter space we sampled from to find good model configurations for each of the RNN types (simple, LSTM, GRU).
  Possible learning rate values were determined on a limited number of preliminary experiments.}
\label{tab:hypers}
\end{table}

We fix the following hyper-parameters for all training runs: we apply data augmentation,
employ stochastic gradient descent with a batch size of 4 and the ADAM update
rule~\cite{kingma_adam_2014}, set all hidden layers to the same (but variable)
size, and stop training if the validation loss does not improve within 15
epochs.

The large number of hyper-parameters prevents us from conducting an exhaustive
search for the optimal architecture. Instead, we use Hyperband
\cite{li_hyperband_2016}, a bandit-based black-box hyper-parameter optimisation scheme, to
find good configurations for each considered RNN type.

In total, we considered 128 configuration for each RNN type (384 different 
models). 
The best RNN setup used one-hot input encoding, skip connections, and $N_h=5$, $D_h=256$, $lr=2.5e^{-4}$.
The best GRU also used one-hot input encoding, but no skip connections, and $N_h=3$, $D_h=512$, $lr=1e^{-3}$.
The best LSTM used a word2vec input encoding with 16 dimensions, skip connections, and $N_h=3$, $D_h=512$, $lr=1e^{-3}$.

In a final step, we further improve the learned models by a fine-tuning stage.
Here, we take the best models found for each RNN type, and re-start training
from the epoch that gave the best results on the validation set, but use
\nicefrac{1}{10} of the original learning rate. As shown in
Table~\ref{tab:results}, this step slightly improves the models.

\section{Results and Discussion}\label{sec:results_and_discussion}

\begin{figure*}[ht!]
  \centering 
  \includegraphics{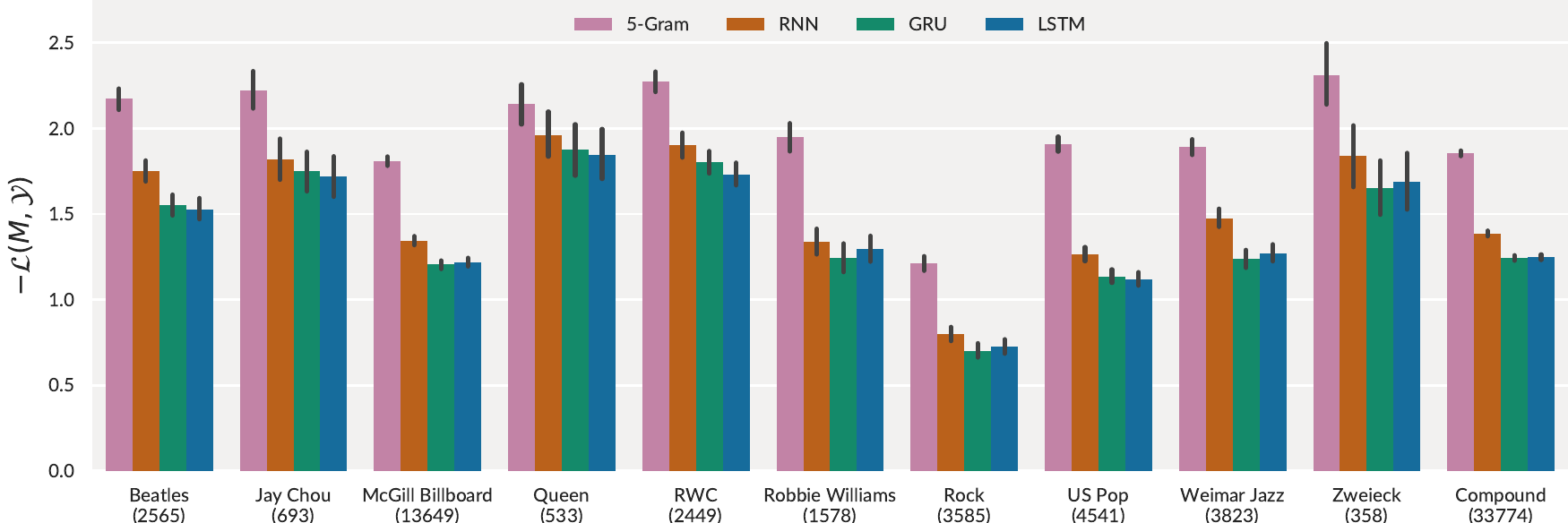}
  \caption{$-\L(M,\Y)$ (lower is better) of the best model for each model
    class on the compound test set, and split up into the individual datasets.
    The numbers in parentheses show the number of chord predictions in each
    set.  Whiskers are 95\% confidence intervals computed using bootstrapping. We
    observe a similar pattern for each set, with the LSTM and GRU performing
    equally on most of the datasets. We also see that chords are easier
    to predict in some datasets (e.g.\ ``Rock''), while more difficult in others,
    (e.g.\ ``RWC'').}\label{fig:overall_results}
\end{figure*}

\begin{figure}[t]
  \centering
  \includegraphics{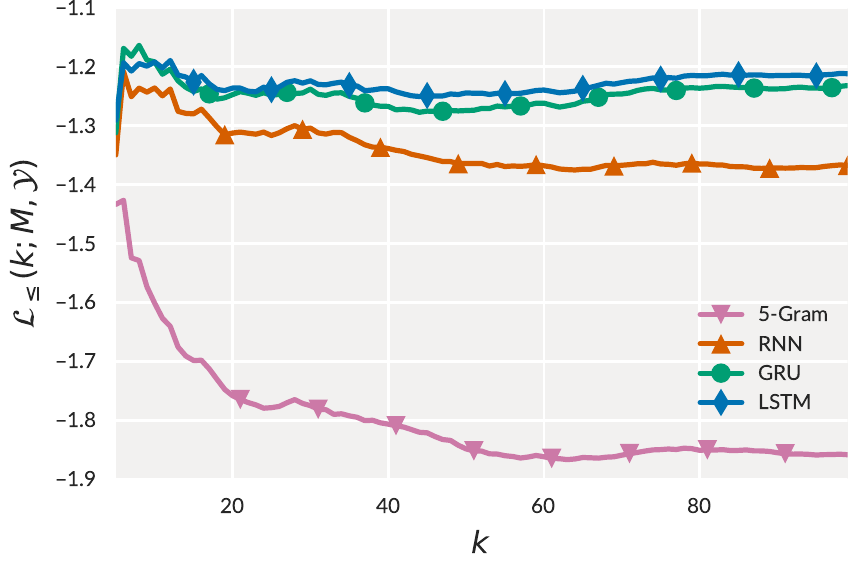}
  \caption{
    Avg.\ cumulative log-probability per chord step $L_\le(k; M, \Y)$. 
    The predictions of the static 5-gram model worsen over time,
    which indicates that chord progressions later in the songs deviate more
    from general patterns than in the beginning. The LSTM and GRU models do not
    suffer from this problem, however, and predictions even improve after chord 40.
    We conjecture that these models better remember previously seen chords
    and patterns, which enables them to automatically adapt as a song progresses.}\label{fig:sequence_results}
\end{figure}

We compare the best model of each RNN type with the 5-gram baseline. To this
end, we compute $\L(M,\Y)$ according to Eq.~\ref{eq:logp} on the test set, and
present the results in Tab.~\ref{tab:results} and
Fig.~\ref{fig:overall_results}.  We see the RNN-based models easily outperform
the 5-gram model, with the LSTM and GRU models performing best. Statistical
significance was determined by a paired t-test with Bonferroni correction.

\begin{table}[]
\sisetup{round-mode=places, round-precision=3}
\centering
\begin{tabular}{@{}lcccc@{}}
\toprule
 & \textbf{5-gram} & \textbf{RNN}    & \textbf{GRU}    & \textbf{LSTM}   \\ \midrule
Val. & \num{1.8300661832473728} & \num{1.4648297553610914}        & \num{1.3283341172995142}         & \num{1.3020447122519005}          \\
Val. (fine-tuned) & --- & \num{1.4174427320839653}         & \num{1.2897256846788903}          & \num{1.2720356053659614}           \\ \midrule
Test & \num{1.8742726483088563} & \num{1.386799} & \textbf{\num{1.2439150}} & \textbf{\num{1.248633}} \\ \bottomrule
\end{tabular}
\caption{$-\L(M,\Y)$ (lower is better) on the validation and test sets for the best model of each model class. All RNN-based models outperform the 5-gram model, with the GRU yielding the highest avg.\ log-probability.
  Statistically equivalent results are marked in bold.}
\label{tab:results}
\end{table}

Recurrent neural networks have the capability to remember
information over time. We thus call them \emph{dynamic} models, compared to the
\emph{static} characteristic of standard models based on N-grams. To
investigate if the RNN-based language models can leverage this
capability, we examine the development of their prediction quality---the 
log-probability of the correct chord---over the progression of songs.

This is a function of both model capacity and song complexity: given a static
model, and repetitive songs that do not change much over time, the
log-probabilities assigned to the chords should remain approximately constant;
however, if e.g.\ chords in interludes tend to deviate more from standard chord
progressions, static models would yield lower log-probabilities in these cases.
On the other hand, if a dynamic model could remember chord progressions
from the past, the predictions would improve over time for very
repetitive songs.

\pagebreak

To evaluate this quantitatively, we first selected from the test set songs that
contained at least 100 chords, which left us with 136 pieces. We then computed
the average cumulative log-probability of chords up to a position in a song as
\begin{align}
  \L_{\le}(k; M, \Y) &= \frac{1}{k|\Y|} \sum_{\y \in \Y} \sum_{\tilde{k}=1}^k \log\left[P_M\left(y_{\tilde{k}} \mid y_1^{\tilde{k}-1}\right)\right]. \label{eq:ts2}
\end{align}
Note that Eq.~\ref{eq:ts2} converges to the objective in Eq.~\ref{eq:logp} with
increasing $k$.

Figure~\ref{fig:sequence_results} presents the results for each model. To our
surprise, the performance of the static 5-gram model drops significantly during
the first 20 chords and continues to fall until around chord 60, after which it
stagnates. This indicates that the chord progressions in the beginning of the
songs are easier to predict by a static model---i.e., more closely
follow the general chord progressions the model learned, while those later in
the songs deviate more from common patterns.

In comparison, the RNN-based models do not suffer during the first 20 chords.
Although they show similar behaviour during the first 20 chords, the LSTM and
GRU models also behave differently later in the songs: both the GRU and the
LSTM improve from around chord 40, whereas the performance of the simple RNN
continues to drop until around chord 60 (similarly to the 5-gram model, but on
a higher level). 

Since predicting chords later in the songs is more difficult than at the
beginning, but the LSTM and the GRU are only negligibly affected by this
(they almost recover to their performance at the beginning), we argue that
both the GRU and the LSTM models are better capable of adapting to the current
song. One explanation for this might be that these models privilege intra-song
statistics during prediction. Thus, rather than learning a global, generic
model trained on the statistics of an entire corpus, we conjecture that these
models also acquire and apply knowledge about the immediate past.

\section{Conclusion}\label{sec:conclusion}

We presented a comprehensive evaluation of chord language models, with a focus
on RNN-based architectures. We discovered the best performing
hyper-parameters, and trained and evaluated the models on a large compound
dataset consisting of various genres. Our results show that
\begin{enumerate*}[label=\arabic*)]
\item all RNN-based models outperform $N$-gram models, 
\item gated RNN cells such as the LSTM cell or the GRU outperform simple RNNs, 
      and 
\item that both LSTM and GRU networks seem to adapt their
      predictions to what they observed in the past.
\end{enumerate*}

We conjecture that to improve the recently stagnant chord recognition results,
we need models with a better understanding of music than has been demonstrated in previous
state-of-the-art systems. This work is a first step towards this
goal. Future research needs to deal with modelling chord durations appropriately,
and with integrating such models with frame-wise acoustic chord classifiers.

\bibliographystyle{IEEEbib}
\bibliography{icassp2018}

\end{document}